\documentclass[letterpaper, 10 pt, conference]{ieeeconf}

\IEEEoverridecommandlockouts
\usepackage[noadjust]{cite}
\usepackage{longtable}
\usepackage{amsmath}
\usepackage{graphicx,epstopdf,amssymb}
\usepackage[dvips]{epsfig}
\usepackage{color}
\usepackage{bbm}
\usepackage{dsfont}
\usepackage{enumerate}
\PassOptionsToPackage{hyphens}{url}
\usepackage{hyperref}
\usepackage{subcaption}
\usepackage[export]{adjustbox}

\newtheorem{theorem}{Theorem}

\newtheorem{remark}{Remark}

\newtheorem{assumption}{Assumption}

\newcommand{\comment}[1]{} 

\newcommand{\carre}{\hfill $\blacksquare$}
\newcommand{\carrew}{\hfill $\square$}

\newcommand{\Rc}[2]{\ensuremath{R^{{C}_{#2}}_{{C}_{#1}}}}
\newcommand{\pc}[2]{\ensuremath{p^{{C}_{#2}}_{{C}_{#1}}}}

\newcommand{\pb}[1]{\ensuremath{p^{A}_{#1}}}
\newcommand{\Rca}[1]{\ensuremath{R^{A}_{{C}_{#1}}}}
\newcommand{\pca}[1]{\ensuremath{p^{A}_{{C}_{#1}}}}
\newcommand{\SO}{\ensuremath{\text{SO}(3)}}
\newcommand{\so}{\ensuremath{\mathfrak{so}(3)}}
\newcommand{\Rn}[1][3]{\ensuremath{\mathds{R}^{#1}}}
\newcommand{\pau}[1]{\ensuremath{\bar p^A_{#1}}}
\newcommand{\pcu}[1]{\ensuremath{\bar p^C_{#1}}}
\newcommand{\grad}{\ensuremath{\text{grad}_1\!}}

\newcommand{\tr}{\ensuremath{\text{tr}}}
\newcommand{\Rorb}[2]{\ensuremath{\check R^{{C}_{#2}}_{{C}_{#1}}}}
\newcommand{\porb}[2]{\ensuremath{\check p^{{C}_{#2}}_{{C}_{#1}}}}

\title{\LARGE A Discrete-Time Attitude Observer on SO(3) for Vision and GPS Fusion
	\vspace{-3mm} 
	}
\author{
Alireza Khosravian, Tat-Jun Chin, Ian Reid, Robert Mahony\\
\thanks{Alireza Khosravian, Tat-Jun Chin, and Ian Reid are with School of Computer Science, University of Adelaide ({\{alireza.khosravian, tat-jun.chin, ian.reid\} @adelaide.edu.au}). Robert Mahony is  with the Research School of Engineering, Australian National University (robert.mahony@anu.edu.au)}
\thanks{This work was supported by the Australian Research Council through the ARC Linkage Project LP140100946, the Centre of Excellence in Robotic Vision CE140100016, and the Laureate Fellowship FL130100102. The authors greatly acknowledge the help and support of Maptek Pty. Ltd., particularly with conducting the experimental studies.}
\vspace{-1.3cm} 
}

\begin{document}

\maketitle

\begin{abstract}
This paper proposes a discrete-time geometric attitude observer for fusing monocular vision with GPS velocity measurements. The observer takes the relative transformations obtained from processing monocular images with any visual odometry algorithm and fuses them with GPS velocity measurements. The objectives of this sensor fusion are twofold; first to mitigate the inherent drift of the attitude estimates of the visual odometry, and second, to estimate the orientation directly with respect to the North-East-Down frame. A key contribution of the paper is to present a rigorous stability analysis showing that the attitude estimates of the observer converge exponentially to the true attitude and to provide a lower bound for the convergence rate of the observer. Through experimental studies, we demonstrate that the observer effectively compensates for the inherent drift of the pure monocular vision based attitude estimation and is able to recover the North-East-Down orientation even if it is initialized with a very large attitude error.
\vspace{-3mm}
\end{abstract}


\IEEEpeerreviewmaketitle

\section{Introduction}
Attitude estimation is the problem of determining the orientation of a vehicle with respect to a known frame of reference. Due to its broad application in robotics, navigation, and control of mechanical systems, this problem has been extensively studied in the past decades \cite{markley2014fundamentals,crassidis2007survey,Mahony08,rehbinder2003pose,grip2012attitude,bonnabel2006non,Vasconcelos08a,zlotnik2015nonlinear,Seo07,Tayebi07McGilvray}. 
High-end IMUs are commonly used for pose estimation in applications, such as mining, that require long term accurate localisation. However, high-end IMUs are usually expensive, heavy, and power hungry. Cameras are attractive alternatives for pose estimation due to their low cost, small size, and high accuracy.
Lots of elegant visual odometry algorithms have been developed in the literature and have been successfully used for localisation of moving robots (see e.g.  \cite{fraundorfer2011visual,fraundorfer2012visual,murTRO2015,engel2013semi,pireIROS15} and the references therein). Monocular visual odometry algorithms rely on matching consecutive images to estimate the relative transformation of camera between those images. This relative transformation is concatenated over time to compute the absolute pose of the camera with respect to the first camera frame. Despite their popularity, there are two potential problems with pure visual odometry in applications that require long term accurate positioning; drift over time, and lack of a global known reference frame.

The concatenated estimate of camera motion based purely on the visual measurements inevitably drifts over time, due to uncertainties in camera intrinsic parameters, aggregation or numerical errors, and measurement noise.
Although attenuated, the drift still exists even if graph based bundle adjustment techniques are used to optimally estimate the relative motion over tens of images \cite{triggs1999bundle}. One way of mitigating the drift is to perform loop closure \cite{angeli2008fast}. However, not all robot paths necessarily contain loops in practice. Moreover, effective loop closure might be difficult in some applications, such as mapping of mine sites, where there is perceptual aliasing in the environment (i.e. several distinct places look similar), or where large moving vehicles, dust, wind, etc. change the appearance of the same location \cite{angeli2008fast}. Also in real time applications, even if effective loop closure could be done, the estimate of robot's path still drifts between two consecutive occurrence of the loop closure.

The lack of a global known reference frame is another problem in pure vision based navigation. Many control objectives require knowledge of the robot's attitude with respect to a fixed known frame of reference (e.g. the North-East-Down (NED) frame) to navigate the robot toward a target point in the environment. Vision based orientation estimates can not be computed with respect to the NED frame as camera alone does not have any information about the NED frame. Although the concatenation of relative transformations yields an estimate of the orientation with respect to the initial camera frame, the transformation of the initial camera frame with respect to the NED frame is still unknown.

Fusing GPS measurements with vision helps mitigating both of the problems discussed above. GPS velocity measurements contain the heading information, that help compensating for the inherent drift of visual odometry estimates. Also, GPS measures the motion directly with respect to a known reference, enabling estimation of the attitude directly with respect to the NED frame. Fusing vision with GPS measurements is particularly of interest in open cut mine sites where RTK GPS is employed to provide centimetre-level position accuracy, albeit with the aid of reference stations. GPS measurements can be quite noisy and unreliable in deep mine pits, hence require fusion with high bandwidth sensors such as IMUs in Inertial Navigation Systems (INS). Fusing the low bandwidth GPS measurements with the high bandwidth visual measurements is a promising solution for long term localisation with a comparable level of accuracy to a high-end INS at a fraction of the cost (see section \ref{sec:experiment}). 

Recursive stochastic filtering techniques, such as (extended) Kalman filtering, are very popular in sensor fusion for navigation \cite{lefferts1982kalman,markley2014fundamentals}. Nevertheless, Kalman filters usually require careful tuning, lack rigorous proof of stability when applied to non-linear systems, and are prone to divergence if not initialised carefully \cite{izadi2015comparison,roberts2003low,Mahony08,zamani2013deterministic,khosravian2010globally,barrau2015invariant}.
To address these issues, geometric nonlinear observer design methods on Lie groups have been developed in the past decade \cite{khosravian2015observers,zamani2013deterministic,lageman2010gradient,bonnabel2009non}. These methods do not provide as rich information as stochastic methods due to the lack of the state covariance. Nevertheless, continuous-time geometric nonlinear attitude observers developed based on these methods are simple to tune, computationally cheap, and have demonstrated good performance in applications where high rate IMUs are present \cite{Mahony08,rehbinder2003pose,grip2012attitude,bonnabel2006non,Vasconcelos08a,zlotnik2015nonlinear,Seo07,Tayebi07McGilvray}.
These methods, however, can not be directly applied to the vision-GPS fusion problem as the underlying kinematics of the this problem is naturally \textit{discrete-time} due to the low rate of camera in most applications. 

Elegant methods for designing discrete-time observers on Lie groups are developed in \cite{barrau2015intrinsic,lee07}. Although suitable for satellite applications, the observer of \cite{lee07} relies on attitude dynamics and the knowledge of external forces applied to the rigid body, hence is not applicable to the problem considered here. The observer of \cite[equ. (18)]{barrau2015intrinsic} assumes availability of at least \textit{two} non-collinear vectorial measurements and provides a novel stability analysis. As we will explain in this paper, however, GPS velocity measurements provide \textit{only a single} vectorial measurement which prevents application of the stability theory of \cite{barrau2015intrinsic} to the particular problem considered here. To the best of our knowledge, there is no discrete-time geometric attitude observer in the literature concerning the problem of fusing vision and GPS velocity measurements (without the aid of IMU).

In this paper, we propose a discrete-time geometric observer on \SO~for fusing monocular visual odometry estimates with GPS velocity measurements. The structure of the proposed observer is inspired by \cite{barrau2015intrinsic} and the proof of stability provided here is an extension to that work based on inspirations from \cite{khosravian2010globally,trumpf2012analysis,grip2012attitude,Seo07}. We show that the estimates of relative translation of camera (provided by a visual odometry algorithm) can be modelled as a vectorial measurement whose associated reference direction is the GPS velocity measurement. Although this is the sole vectorial measurement available in this application, we show that the attitude can be asymptotically estimated if the direction of the velocity of vehicle varies sufficiently with time (see Assumption \ref{as:persistancy}). A key contribution of this paper is to provide a rigorous stability analysis yielding a deep understanding of the underlying fundamental performance limits of the vision and GPS fusion problem in terms of the convergence rate of the observer. Our fusion method is independent of the choice of the visual odometry algorithm, giving the users an extra freedom to apply any algorithm that suites their particular application. We demonstrate the application of the developed theory by experimental studies showing excellent performance of the proposed observer in compensating for the drift in the visual odometry and recovering the orientation of vehicle with respect to the NED frame.

The structure of paper is as follows. In Sections \ref{sec:problem:formulation} and \ref{sec:simplified:problem:formulation} formulate, for the first time, the attitude estimation problem using vision and GPS velocity fusion as a discrete-time geometric observer design problem on \SO  with an associated vectorial measurement belonging to the unit sphere. The attitude observer is proposed in Section \ref{sec:observer:design} and the stability analysis is performed in Section \ref{sec:stability} (see Theorem \ref{theo:stability}). We provide simulation studies as well as experimental results in Sections \ref{sec:simulation} and \ref{sec:experiment}, respectively, and we finish the paper with conclusions in Section \ref{sec:conclusion}.

\section{Problem formulation} \label{sec:problem:formulation}
Consider a vehicle equipped with a vision system and a GPS receiver
moving in an environment with an unknown map. Assume that the GPS receiver and the camera
are rigidly connected to the vehicle. Denote the North-East-Down (NED) reference frame (i.e. inertial frame) by $\{A\}$ and suppose that the reference frame $\{C\}$ is attached to the camera and moves with it. For simplicity, we refer to the camera reference frame when the $k$-th image was captured as the $k$-th camera frame. Denote the rotation of the $i$-th camera frame with respect to (wrt.) the $j$-th camera frame by $\Rc{i}{j}\in\SO$. By matching corresponding feature points in two consecutive images $k$ and $k+1$, visual odometry algorithms are capable of computing the relative rotation between the camera frames of those two images \cite{fraundorfer2011visual,fraundorfer2012visual}
We assume that \Rc{k+1}{k} is provided by a visual odometry algorithm, and we treat this variable as a known measurement. One can obtain the rotation of the $(k+1)$-th camera frame wrt. the initial camera frame $C_0$ by concatenating previous relative rotations, as follows.
\begin{align}
\label{eq:Rc:dyn} \Rc{k+1}{0} &=\Rc{k}{0} \Rc{k+1}{k},~~~~~&&\Rc{0}{0}=I_3.
\end{align}
Visual odometry techniques are also able to compute the relative translation of the consecutive camera frames. For monocular cameras, however, such a relative translation is valid only up to a constant scale since the depth can not be measured in the monocular setup \cite{hartley2003multiple}. Denote the translation of the $i$-th camera frame wrt. the $j$-th camera frame (and expressed in the $j$-th camera frame) by $\pc{i}{j}\in\Rn$. Assuming that the camera and GPS receiver are placed sufficiently close together\footnote{It is possible to extend the derivations of this paper to the case where the camera and the GPS receiver are not placed close together. We consider that as a possible future work.}, it is shown in Appendix \ref{sec:appendix:GPS2cam} that 
\begin{align}
\label{eq:GPS:difference:model} \pc{k+1}{k}= \frac{1}{d}R_k^\top (\pb{k+1}-\pb{k})
\end{align}
where $\pb{k+1}\in \Rn$ is the $k$-th camera position (wrt. and expressed in the NED frame), \pc{k+1}{k} is the relative translation of the $k$-th camera frame wrt. the $(k+1)$-th camera frames (expressed in the $k$-th camera frame) computed by the visual odometry algorithm and treated here as a known measurement, and $R_k \in \SO$ is the rotation of the $k$-th camera frame wrt. the NED frame. This rotation is related to $\Rc{k}{0}$ via $R_k= \Rca{0} \Rc{k}{0}$ where $\Rca{0} \in \SO$ is the unknown initial rotation of the camera frame wrt. the NED frame. One can use the GPS position measurements to compute $\pb{k+1}-\pb{k}$ in (\ref{eq:GPS:difference:model}), nevertheless, numerical differencing the noisy GPS position measurements further increases the level of noise. For this reason, it is recommended to use the GPS velocity measurements to approximate \pb{k+1}-\pb{k} as follows\footnote{Velocity measurements are computed in many GPS chips from the Doppler shift of the pseudo-range signals, yielding far more accurate velocity measurements compared to numerically differencing the range based position measurements. This accuracy is further enhanced if time-differenced carrier phase (TDCP) based GPS is used.}.
\begin{align}
\pb{k+1}-\pb{k} = \int_{t_k}^{t_{k+1}}\!\!\! v^A(s) \text{d}s \approx (t_{k+1}-t_k) \bar{v}_k^A.
\end{align}
If the  vehicle's velocity is approximately constant from $t_k$ to $t_{k+1}$, we simply use $\bar{v}_k^A=v_k^A \in \Rn$ where $v_k^A$ is the GPS velocity measurement at time $t_{k}$. Otherwise, if the velocity changes linearly with time from $t_k$ to $t_{k+1}$, we use the approximation $\bar{v}_k^A = 0.5 (v_{k+1}^A+v_k^A)$.

The aim of this paper is to estimate $R_k$ using the discrete-time dynamics model (\ref{eq:Rc:dyn}) and the output measurement model (\ref{eq:GPS:difference:model}). That is to fuse the outputs of visual odometry algorithm \Rc{k+1}{k} and \pc{k+1}{k} with the GPS velocity measurement $v_k$ (or equivalently the difference of consecutive position measurements if the velocity is unavailable). There are several challenges in such an estimation problem, some of which have been listed below.

\begin{itemize}
	\item As opposed to classical linear estimation problem where the state space is a vector space, here the underlying system (\ref{eq:Rc:dyn}) evolves on the manifold \SO.
	\item The underlying dynamics is discrete-time, which prevents direct application of previous continuous-time observer design techniques on Lie groups \cite{khosravian2015observers,zamani2013deterministic,lageman2010gradient,bonnabel2009non}.
	\item The scale $d$ of the translation measurements provided by the visual odometry algorithm is unknown.
	\item The GPS measurements do not immediately provide information about the camera rotation. More specifically, the GPS measurements are related to the camera orientation via (\ref{eq:GPS:difference:model}) where the initial camera orientation \Rca{0} is unknown. Also, since (\ref{eq:GPS:difference:model}) only provides a single $3$D vectorial measurement, the rotation is not instantaneously observable.
\end{itemize}

Note that we do not consider stochastic measurement noise in our modeling as our design framework is based on the deterministic geometric setup \cite{khosravian2015observers,zamani2013deterministic,lageman2010gradient,bonnabel2009non}.

\section{Vision-GPS fusion as a geometric observer design problem} \label{sec:simplified:problem:formulation}
In this section, we transform the underlying system kinematics (\ref{eq:Rc:dyn}) and the output model (\ref{eq:GPS:difference:model}) to a form more suitable for observer design purpose. We normalize the measurements involved in (\ref{eq:GPS:difference:model}) to eliminate the unknown scale $d$. Defining
\begin{align}
\label{eq:pau} \pau{k}&:=\frac{\pb{k+1}-\pb{k}}{\|\pb{k+1}-\pb{k}\|} \approx \frac{\bar{v}_k^A}{\|\bar{v}_k^A\|},\\
\label{eq:pcu} \pcu{k}&:=\frac{\pc{k+1}{k}}{\|\pc{k+1}{k}\|},
\end{align}
we have
\begin{align}
\label{eq:output:model} \pcu{k} =  R_k^\top \pau{k}.
\end{align}
The variables $\pcu{k}$ and $\pau{k}$ are unit vectors expressed in the camera frame and the NED frame, respectively. The normalization is particularly important since the scale of the visual odometry estimates drifts over time. The normalization not only isolates the effect of the scale drift, but also eliminates the need for knowledge of the velocity time stamp $t_k$. Equation (\ref{eq:output:model}) formulates the output measurement model of the system as a right action of \SO (the state space) on $\text{S}^2$ (the output manifold) where \pau{k} represent a \textit{time-varying reference output} in this context \cite{khosravian2015observers,mahony2013observers,khosravian2010globally,trumpf2012analysis}. This model directly relates the measurements $\pcu{k}$ and $\pau{k}$ to the unknown state $R_k$ and is used in the observer design.

Multiplying the sides of (\ref{eq:Rc:dyn}) by the constant rotation $\Rca{0}$ from the left and recalling $R_k=\Rca{0} \Rc{k}{0}$, we have
\begin{align}
\label{eq:Rk:dyn} R_{k+1} = R_k \Rc{k+1}{k},~~~~~R_0=\Rca{0}.
\end{align}
Now, the observer design problem is simplified to estimate $R_k$ governing the dynamics (\ref{eq:Rk:dyn}) using the measurement of \Rc{k+1}{k} (as the input of the dynamics (\ref{eq:Rk:dyn})) and the vectorial measurements \pcu{k} and \pau{k} modeled via (\ref{eq:output:model}). The main advantage of formulating the problem as above compared to Section \ref{sec:problem:formulation} is that the initial camera rotation $\Rca{0}$ now appear as an unknown initial condition of the estimation problem rather than as unknown variables in the system equation.

\section{Observer design on \SO} \label{sec:observer:design}
The underlying kinematics of the system (\ref{eq:Rk:dyn}) is left-invariant and the considered output model (\ref{eq:output:model}) is right-invariant (in the sense of \cite{khosravian2015observers,mahony2013observers}). Inspired by the novel work of \cite{barrau2015intrinsic}, we consider the following structure for our observer.
\begin{align}
\label{eq:obs:grad:structure} \hat R_{k+1} = \hat R_k \exp(-\hat R_k^\top \grad\phi(\hat R_k,\pcu{k})) \Rc{k+1}{k},
\end{align}
with $\hat R_0=I_3$, where $\phi : \SO \times \text{S}^2 \to \mathds{R}^+$ is a right-invariant cost function (in the sense of \cite{khosravian2015observers}) and $\grad\phi(\hat R_k,\pcu{k}) \in T_{\hat R_k}\SO$ is the gradient of $\phi(.,\pcu{k}): \SO \to \Rn[^+]$ wrt. a right-invariant Riemannian metric (see \cite[Proposition 6.1.]{khosravian2015observers}). The motivation behind the observer structure (\ref{eq:obs:grad:structure}) is that it generates an autonomous gradient-like estimation error dynamics if the reference output is constant \cite{barrau2015intrinsic}. We naturally choose the simple cost function
\begin{align}
\label{eq:cost} \phi(\hat R_k,\pcu{k})=(\hat R_k \pcu{k} - \pau{k})^\top L (\hat R_k \pcu{k} - \pau{k})
\end{align}
where $L \in \Rn[3\times3]$ is a constant positive definite matrix. Observe that $\hat R_k = R_k$ implies $\phi=0$. Employing standard differential geometry derivations, the gradient of $\phi$ wrt. the induced Riemannian metric $\langle \Omega_1 R, \Omega_2 R \rangle_R = \tr(\Omega_2^\top \Omega_1)$ is given by \cite[Section VI]{khosravian2013bias}
\begin{align}
\label{eq:grad:phi} \grad\phi(\hat R_k,\pcu{k}) = -\Big(\!\!\big(L (\hat R_k \pcu{k}-\pau{k})\!\big)_{\!\times} \hat R_k \pcu{k}\Big)_{\!\times} \hat R_k.
\end{align}
where $(.)_\times$ denotes the isomorphism from $\Rn[3]$ to $\so$ such that $a_\times b$ yields the vector cross product $a \times b$ for any $a,b \in \Rn[3]$. Using (\ref{eq:obs:grad:structure}) and (\ref{eq:grad:phi}), and employing the property $\exp(\hat R^\top a_\times \hat R) = \hat R^\top \exp(a_\times) \hat R$, we propose the observer
\begin{align}
\label{eq:obs:Rk} \hat R_{k+\!1} \!\!=\! \exp\!\Big(\!\!\Big(\!\!\big(L (\hat R_k \pcu{k}-\pau{k})\!\big)_{\!\times} \hat R_k \pcu{k}\Big)_{\!\times}\!\Big) \hat R_k \Rc{k+\!1}{k},~\hat R_0\!\!=\!\!I_3.
\end{align}

\begin{remark} \label{rem:sampling:rate}
The observer (\ref{eq:obs:Rk}) assumes GPS measurements are synchronized with images. In practice, the sampling rate of GPS is usually lower than that of the camera. In this case, one can introduce the auxiliary state $\hat R_k^+$ and break the observer (\ref{eq:obs:Rk}) into two stages; prediction and update. In the prediction stage, the auxiliary state is updated by $\hat R_{k+1}^+= \hat R_k \Rc{k+\!1}{k}$ each time a new relative orientation $\Rc{k+\!1}{k}$ is provided by visual odometry. In the update stage, the state $\hat R$ is updated via $\hat R_{k+\!1} = \exp\!\Big(\!\!\Big(\!\!\big(L (\hat R_k^+ \pcu{k}-\pau{k})\!\big)_{\!\times} \hat R_k^+ \pcu{k}\Big)_{\!\times}\!\Big) \hat R_k^+ $ when GPS provides a new measurement. It is possible to show that the stability results of Section \ref{sec:stability} still hold when the two-stage implementation of the observer is employed.
\end{remark}

\section{Stability analysis} \label{sec:stability}
Although the structure of the proposed observer (\ref{eq:obs:Rk}) is inspired by the novel work of \cite{barrau2015intrinsic}, the stability analysis presented in \cite[Proposition 4]{barrau2015intrinsic} does not apply here as the cost function (\ref{eq:cost}) does not have a unique global minimum. Instead, we provide a new stability analysis by extending the results of \cite{khosravian2010globally,trumpf2012analysis}. Consider the attitude estimation error.
\begin{align}
\label{eq:Ek} E_k=\hat R_k R_k^\top.
\end{align}
To prove $\hat R_k \to R_k$, we should show $E_k \to I_3$. Using (\ref{eq:Rk:dyn}), (\ref{eq:obs:Rk}), (\ref{eq:output:model}), and (\ref{eq:Ek}), the dynamics of $E_k$ are given by
\begin{align}
  \nonumber E_{k+1} &\!=\! \hat R_{k+1} R_{k+1}^\top \!=\!\exp\!\Big(\!\!\Big(\!\!\big(L (\hat R_k \pcu{k}\!-\!\pau{k})\!\big)_{\!\times} \!\hat R_k \pcu{k}\Big)_{\!\times}\Big) \hat R_k R_k^\top\\
\label{eq:E:dy}  &=\exp\!\Big(\Big(\!\!\big(L (E_k \pau{k}-\pau{k})\!\big)_{\!\times} E_k \pau{k}\Big)_{\!\times}\Big) E_k.
\end{align}
One can verify that $E_k=I_3$ is an equilibrium point of the error dynamics (\ref{eq:E:dy}). If $\pau{k}$ were constant and if we had two or more vector measurements, the elegant stability analysis of \cite[Proposition 4]{barrau2015intrinsic} would be applied. However, only one vector measurement is available here. Since $\pau{k}$ is time-varying, the stability analysis is tedious even in the continuous-time case \cite{khosravian2010globally,trumpf2012analysis}. Inspired by the continuous-time case, here we propose a novel stability analysis for the discrete-time case. The following assumption formally formulates the time-varying property of \pau{k}.

\begin{assumption}[Persistency of excitation] \label{as:persistancy}
	There exist a positive integer $T$ and a constant $\beta > 0$ such that
	\begin{align}
	\label{eq:persistency}	\frac{1}{T+1} \sum_{i=k}^{k+T} (I_3 -\pau{i} \pau{i}{}^\top) \ge \beta I_3,
	\end{align}
	for all non-negative integers $k$. \carrew
\end{assumption}

Assumption \ref{as:persistancy} could be written in other equivalent forms, see \cite[Lemma 1]{khosravian2010globally}. This assumption is satisfied if \pau{k} varies with time, i.e. if the vehicle's direction of motion is not constant all the time (see \cite[Lemma 1]{khosravian2010globally}). The only case where this assumption does not hold is when the vehicle's trajectory is a perfectly straight line. It is worth mentioning that in the closely related problem of velocity-aided attitude estimation where an IMU is present instead of a camera, a similar condition as (\ref{eq:persistency}) is required unless an extra sensor (such as magnetometer) is available \cite{hua2010attitude}. The more the direction of the motion of the vehicle changes, the larger $\beta$ gets. Nevertheless, since \pau{k} is a unit vector, the largest singular value of the matrix $I_3 -\pau{i} \pau{i}{}^\top$ is one and we have $\beta < 1$ \cite{khosravian2010globally}. The following theorem summarizes the stability of the error dynamics (\ref{eq:E:dy}). To simplify presentation of the results, we consider the observer gain $L=l I_3$ where $l$ is a scalar. It is straight-forward to extend the results to a more general positive definite gain matrix.  

\begin{theorem} \label{theo:stability}
	Consider the observer (\ref{eq:obs:Rk}) for the system (\ref{eq:Rk:dyn}) with the output (\ref{eq:pau})-(\ref{eq:output:model}). Suppose that Assumption \ref{as:persistancy} holds. Assume moreover that $L=l I_3$ where $0 < l < 2$. Then, the error dynamics (\ref{eq:E:dy}) is locally exponentially stable to $E_k=I_3$. Moreover, there exists a compact region around $E_k=I$ in which
	\begin{align}
	\label{eq:exponential:bound}	\|I_3 - E_k\|_F \le \|I_3 - E_0\|_F \alpha^k,
	\end{align}
	where $0<\alpha:=\Big(1- \frac{\beta (T+1) (2l-l^2)}{2 + l^2 T (T+1)} \Big)^{\frac{1}{2(1+T)}} < 1$ indicates a lower bound for the convergence rate and $\|I_3 - E_k\|_F^2=\tr((I_3-E_k)^\top (I_3-E_k))=2\tr(I_3-E_k)$ is the induced Frobenius norm on \SO.\carrew
\end{theorem}

\textit{Proof of Theorem \ref{theo:stability}:} the proof is mainly inspired by \cite[Section V]{khosravian2010globally} which investigates the continuous-time scenario. There are subtle technical differences in the discrete-time case that prevent direct application of the continuous-time case. For this reason, here we provide a stability proof for the discrete-time case in detail. The first order approximation of the error $E_k$ around the identity is given by $E_k \approx I+{\epsilon_k}_\times$ where $\epsilon_k \in \Rn$. Substituting for this approximation into the error dynamics (\ref{eq:E:dy}) and resorting to Appendix \ref{sec:appendix:linearization}, we obtain the linearized error dynamics
\begin{align}
\label{eq:err:dy:linear}	\epsilon_{k+1} = \epsilon_{k} - l P_k \epsilon_k,
\end{align}
where
\begin{align}
\label{eq:P} P_k:= -\pau{k}{}_\times \pau{k}{}_\times=I_3 - \pau{k} \pau{k}{}^\top	
\end{align}
is a time-varying positive semi-definite matrix whose rank is two. The Assumption \ref{as:persistancy} imposes a persistency of excitation condition on $P_k$ which can increase the rank of its summation over time to three. Consider the following Lyapunov candidate.
\begin{align}
\label{eq:L} \mathcal{L}_k = \epsilon_k^\top \epsilon_k.
\end{align}   
Using (\ref{eq:err:dy:linear}) and noting that $P_k^\top = P_k$ and $P_k^2=P_k$ (due to (\ref{eq:P}) and the fact that $\pau{k}$ has a unit norm), we have
\begin{align}
	\mathcal{L}_{k+1} - \mathcal{L}_{k} &= (\epsilon_{k} - l P_k \epsilon_k)^\top (\epsilon_{k} - l P_k \epsilon_k) -  \epsilon_k^\top \epsilon_k\\
\label{eq:dL}	&= - \gamma \epsilon_k^\top P_k \epsilon_k.
\end{align}
where $\gamma :=2 l-l^2$. Since $0 < l < 2$, we have $0 < \gamma < 1$. Since $P_k$ is positive semi-definite, we have $\mathcal{L}_{k+1} - \mathcal{L}_{k} \le 0$ which indicates that the Lyapunov candidate is non-increasing along the system trajectories. Summing the sides of (\ref{eq:dL}) from $k$ to $k+T$ yields
\begin{align}
\label{eq:lyapunov:integral}	\mathcal{L}_{k+1+T}-\mathcal{L}_{k} = - \gamma \sum_{i=k}^{k+T} \epsilon_i^\top P_i \epsilon_i
\end{align}
Since $P_i \ge 0$, there exist $N_i \in\Rn[2 \times 3]$ (with rank 2) such that $P_i = N_i^\top N_i$. Choose $a=N_i \epsilon_k$ and $b=N_i(\epsilon_i-\epsilon_k)$ and employ the property $(a+b)^\top (a+b) \ge \frac{1}{2} a^\top a - b^\top b$ to obtain
\begin{align}
\label{eq:inequality:tmp1}	\sum_{i=k}^{k+T} \epsilon_i^\top P_i \epsilon_i \ge \frac{1}{2} \sum_{i=k}^{k+T} \epsilon_k^\top P_i \epsilon_k - \sum_{i=k}^{k+T} (\epsilon_i^\top - \epsilon_k^\top) P_i (\epsilon_i - \epsilon_k).
\end{align}
Using the derivations of Appendix \ref{sec:appendix:integ:inequality}, we have
\begin{align}
\label{eq:inequality:int}	\sum_{i=k}^{k+T} (\epsilon_i^\top - \epsilon_k^\top) P_i (\epsilon_i - \epsilon_k) \le \frac{l^2 T(T+1)}{2}  \sum_{i=k}^{k+T} \epsilon_i^\top P_i \epsilon_i
\end{align}
Also, (\ref{eq:persistency}) and (\ref{eq:L}) yield
\begin{align}
\label{eq:inequality:pe}	\sum_{i=k}^{k+T} \epsilon_k^\top P_i \epsilon_k \ge  \beta (T+1) \epsilon_k^\top \epsilon_k = \beta (T+1) \mathcal{L}_k
\end{align}
Substituting for (\ref{eq:inequality:int}) and (\ref{eq:inequality:pe}) into (\ref{eq:inequality:tmp1}), we obtain
\begin{align}
\label{eq:inequality:tmp2}	\sum_{i=k}^{k+T} \epsilon_i^\top P_i \epsilon_i \ge \frac{\beta (T+1)}{2}   \mathcal{L}_k - \frac{l^2 T (T+1)}{2}  \sum_{i=k}^{k+T} \epsilon_i^\top P_i \epsilon_i,
\end{align}
which implies $\sum_{i=k}^{k+T} \epsilon_i^\top P_i \epsilon_i \ge=\frac{\beta (T+1)}{2 + l^2 T (T+1)} \mathcal{L}_k$. Substituting this into (\ref{eq:lyapunov:integral}) yields
\begin{align}
\label{eq:Lyapunov:reduce} \mathcal{L}_{k+1+T} \le  \bar \alpha \mathcal{L}_k.
\end{align}
where $\bar \alpha:=1- \frac{\beta (T+1) (2l-l^2)}{2 + l^2 T (T+1)}$. Recalling that $0 <l < 2$ and $T \ge 1$, it is straight-forward to show that $1-\frac{\sqrt{2}}{2} \beta <\bar \alpha < 1$. Employing (\ref{eq:P}) and recalling that \pau{k} is a unit vector, we conclude that the largest singular value of $P_k$ is one. This, together with (\ref{eq:persistency}) implies that $0<\beta \le 1$. Hence, $0<\frac{\sqrt{2}}{2} <\bar \alpha < 1$. Taking $k= n(1+T)$, where $n$ is a non-negative integer, and successively employing (\ref{eq:Lyapunov:reduce}) yields
\begin{align}
	\mathcal{L}_k \le \mathcal{L}_{n(1+T)} \le \bar \alpha^n \mathcal{L}_0,~~~\forall k\ge n(1+T)
\end{align}
Hence, $\mathcal{L}_k$ decays exponentially to zero as $n$ goes to infinity. Substituting $\mathcal{L}_k$ for (\ref{eq:L}) and considering $n=\frac{k}{1+T}$ yields
\begin{align}
\label{eq:e:exponential:bound} \|\epsilon_k\|\le \|\epsilon_0\| \alpha^k
\end{align}
where $\alpha$ is defined in the statement of Theorem \ref{theo:stability}. Consequently, the equilibrium point $\epsilon_k=0$ of the linearized error dynamics (\ref{eq:err:dy:linear}) is uniformly globally exponentially stable \cite{kalman1960control}. Hence, by extending the results of \cite[Theorem 4.15]{Khalil} to the discrete-time case according to the guidelines of \cite{kalman1960control}, one concludes that the equilibrium point $E_k=I$ of the error dynamics (\ref{eq:E:dy}) is uniformly locally exponentially stable. The exponential bound (\ref{eq:exponential:bound}) follows from (\ref{eq:e:exponential:bound}) directly.\carre

By Theorem \ref{theo:stability}, the convergence rate of the observer depends on the value of $\beta$ which in turn depends on the motion of the vehicle (equ. (\ref{eq:persistency})). Having an analytical solution for the convergence rate (provide by Theorem \ref{theo:stability}) makes it possible to obtain the value of the observer gain $l$ that yields the fastest convergence rate. This can be done by considering a moving horizon of time and numerically computing $T$ and $\beta$ satisfying (\ref{eq:persistency}) for that horizon. Given $\beta$ and $T$, one can analytical compute the value of the observer gain $l$ that maximizes the convergence rate $\alpha$. This yields a time-varying gain as $T$ and $\beta$ should be updates as the time horizon changes. It is straight-forward to extend the stability results of Theorem \ref{theo:stability} to a time-varying gain.

\section{Simulation results} \label{sec:simulation}
\begin{figure} 
\centering
\includegraphics[width=9cm]{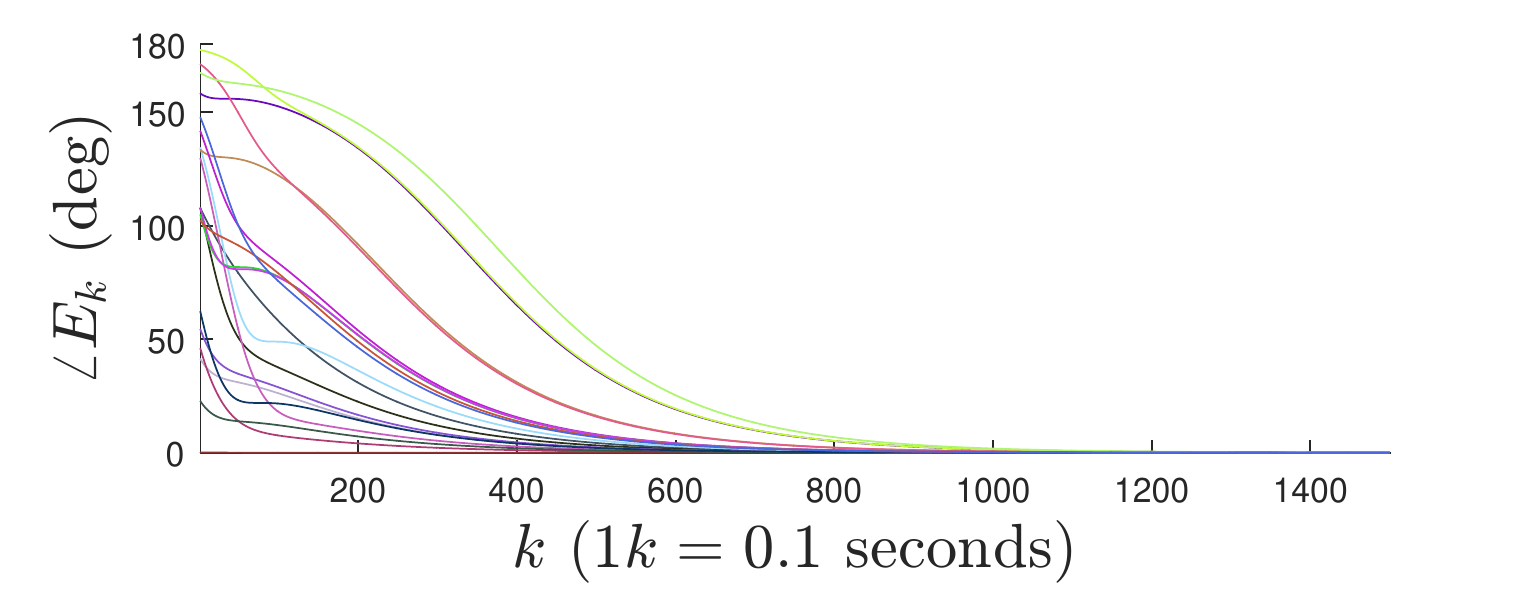}
\vspace{-0.7cm}

\caption{Attitude estimation error of the observer (\ref{eq:obs:Rk}) with different initialization error. Note that $180$ (deg) indicates the maximum possible error.} \label{fig:montecarlo}
\vspace{-0.5cm}
\end{figure}

Theorem \ref{theo:stability} proves local stability of the error dynamics around the identity, but it does not provide an estimate of the domain of attraction. As the initial orientation of the camera wrt. to the NED frame is unknown, one cannot always choose the initial condition of the observer (\ref{eq:obs:Rk}) close to the actual value of \Rca{0}. The aim of this section is to examine the domain of attraction of the error dynamics (\ref{eq:E:dy}). Consider a camera moving with a constant velocity of $2 \pi \approx 6.3$ (m/s) in a circular trajectory with the radius of $50$ (m) in the North-East plane. The NED frame is set to be at the centre of the circle. The camera starts its motion from the far North part of the path and moves clockwise to East. Assuming the sampling interval of $0.1$ seconds, the corresponding relative motions \Rc{k+1}{k} and \pc{k+1}{k} are computed to model the output of visual odometry and passed to the observer (\ref{eq:obs:Rk}) together with the linear velocity of the camera wrt. the NED frame. We neither model camera images nor we add noise to the measurements as the aim of this section is to only investigate the effect of initial condition on the attitude estimation error $\angle E_k = \text{acos}(1-\frac{1}{4}\|I-E_k\|_F^2)$\footnote{The error $\angle E_k$ corresponds to the angle of rotation in the angle-axis decomposition of $E_k$ which models how far $\hat R_k$ is from $R_k$ \cite{lageman2010gradient,khosravian2010globally}.}. We perform $20$ simulations where in each simulation the observer is initialised with a random rotation corresponding to a random axis of rotation in a unit spare and a random angle of rotation between $0$ and $179$ degrees. Fig. \ref{fig:montecarlo} shows the results of the simulations, demonstrating that the attitude estimation error converges to zero even when a very large initial estimation error is chosen.

\section{Experimental results} \label{sec:experiment}
\begin{figure}[!t]
	\centering
	\includegraphics[width=8.5cm]{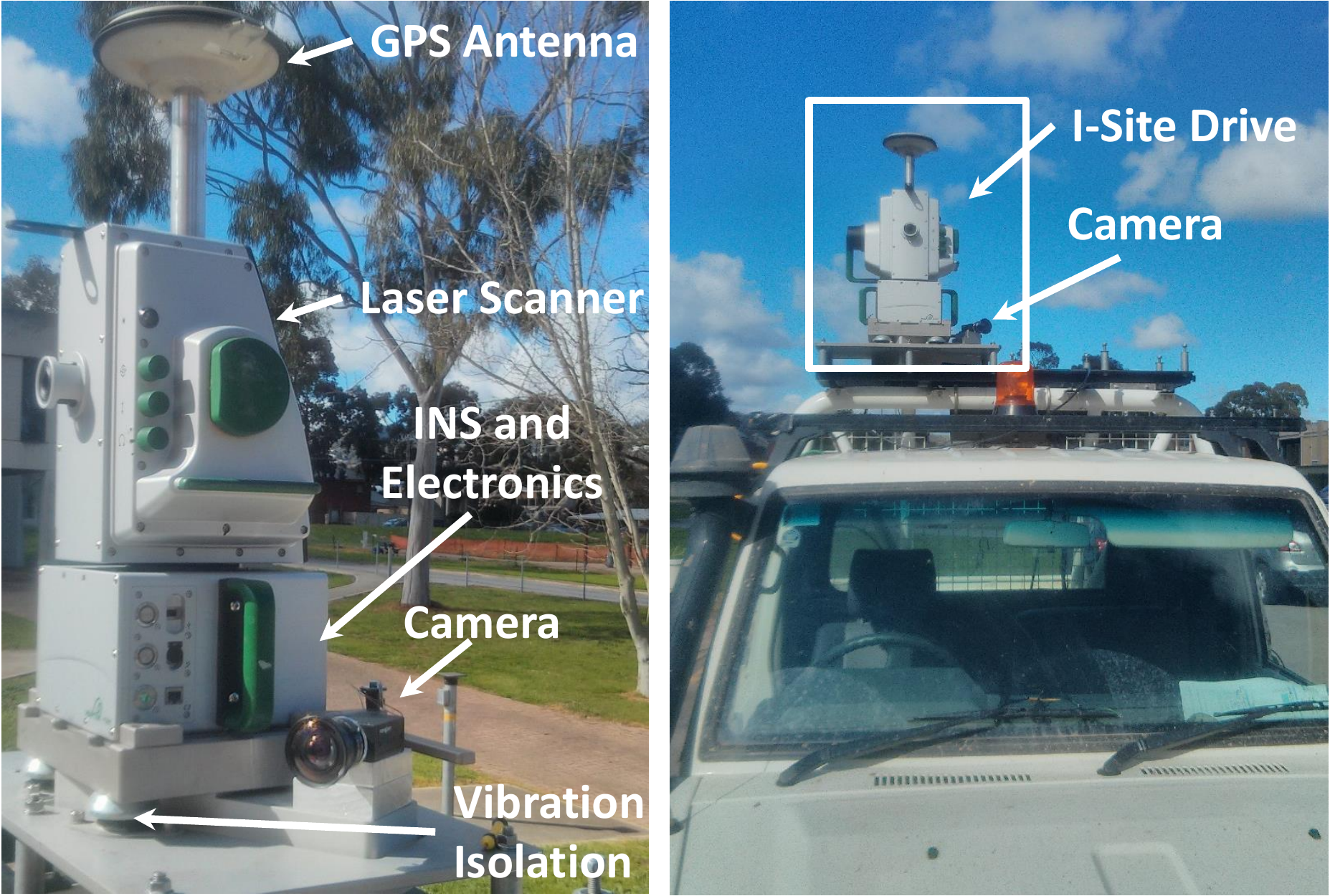}
	\caption{Right: the experimental setup mounted on a truck. Left: enlarged photo of the experimental setup showing its components.} \label{fig:experiment:setup}
%
	\includegraphics[width=9cm]{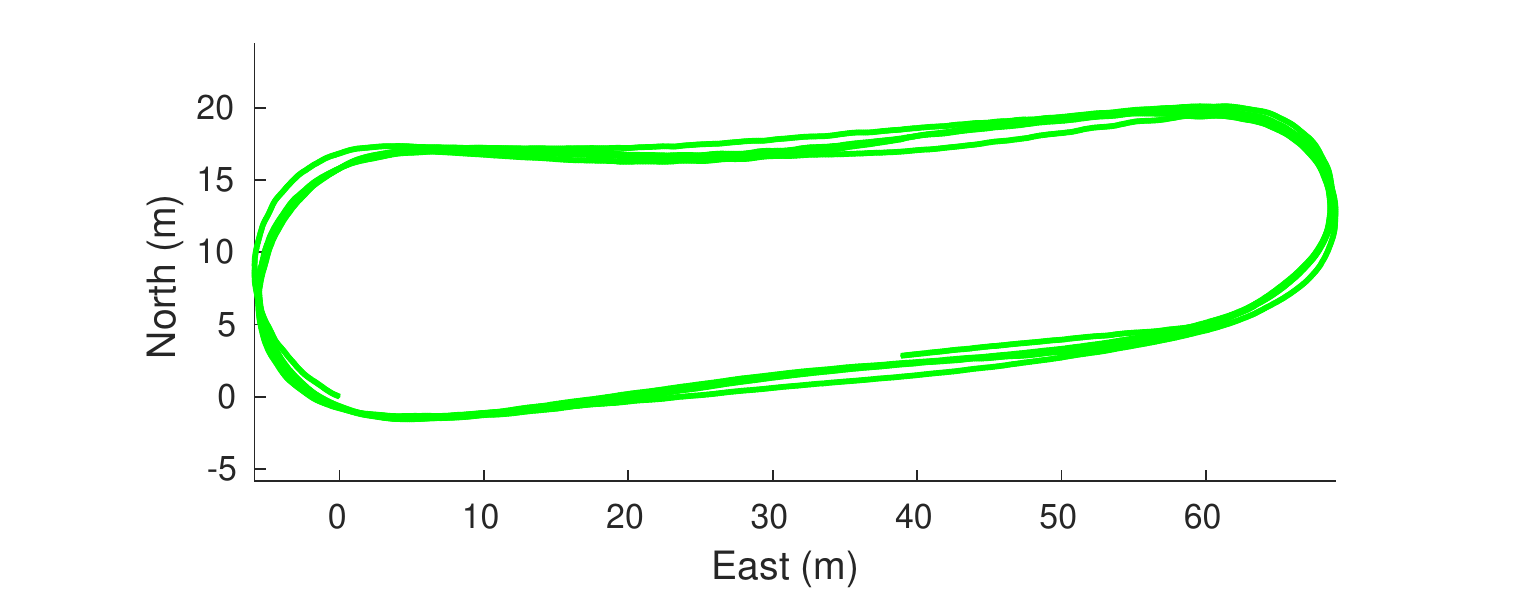}
	\caption{Top view of the ground truth trajectory of the vehicle. The vehicle starts at the origin and moves clockwise.} \label{fig:trajectory}
	\vspace{-5mm}
\end{figure}

In this section, we demonstrate the performance of the proposed observer (\ref{eq:obs:Rk}) with real world experiments. The experiments have been done in collaboration with Maptek Pty Ltd\footnote{\url{http://www.maptek.com/}}. The experimental setup consists of Maptek's I-Site Drive system \footnote{\url{http://www.maptek.com/products/i-site/i-site_drive.html}} together with a Pointgrey Grasshopper3 GS3-U3-23S6C-C camera\footnote{\url{https://www.ptgrey.com/grasshopper3-23-mp-color-usb3-vision-sony-pregius-imx174}} mounted on a truck (see Fig \ref{fig:experiment:setup}). The I-Site Drive includes the NovAtel SPAN-CPT\footnote{\url{http://www.novatel.com/products/span-gnss-inertial-systems/span-combined-systems/span-cpt/}} high precision Inertial Navigation System (INS) consist of a RTK GPS and an inertial measurement unit, providing accurate estimates of the 6DoF transformation of the I-Site Drive system wrt. the inertial frame which is used as the ground truth $(R_k,p_k)$ in our experiments. The extrinsic camera to I-Site Drive calibration transformation have been carefully computed using a hand-eye calibration method that enables computing ground truth transformation of the camera wrt. the NED frame \cite{horaud1995hand}. The camera is configured to capture $1920 \times 1200$ pixels images at approximately $22$ fps. The camera is hardware synced to the I-Site Drive system to provide precise time stamping of measurements for accurate evaluation of the results. In our experiments, we use ORB-SLAM \cite{murTRO2015} as the visual odometry algorithm. ORB-SLAM is a state-of-the-art large scale simultaneous localisation and mapping software. Direct outputs of ORB-SLAM are estimates of the translation of the $k$-th camera frame wrt. the initial camera frame. Given two consecutive transformation estimates $(\Rorb{k}{0},\porb{k}{0})$ and $(\Rorb{k+1}{0},\porb{k+1}{0})$ as direct outputs of ORB-SLAM, we compute $\Rc{k+1}{k}=\Rorb{k}{0}{\!}^\top \Rorb{k+1}{0}$ and $\pc{k+1}{k}=\Rorb{k}{0}{}^\top (\porb{k+1}{0}-\porb{k}{0})$ (recalling (\ref{eq:Rc:dyn}) and (\ref{eq:pc:dyn})) and we use them in the observer (\ref{eq:obs:Rk}). For our experiments, we disable the loop closure thread of ORB-SLAM to observe the inherent drift of the pure visual odometry\footnote{We recall that effective loop closure is not possible in some mining applications due to the restrictions of vehicle's path (not necessarily containing loops), high perceptual aliasing, large moving vehicles, and change in the appearance of the environment due to dust, wind, etc.}.

\begin{figure}[!t]
\includegraphics[width=9cm]{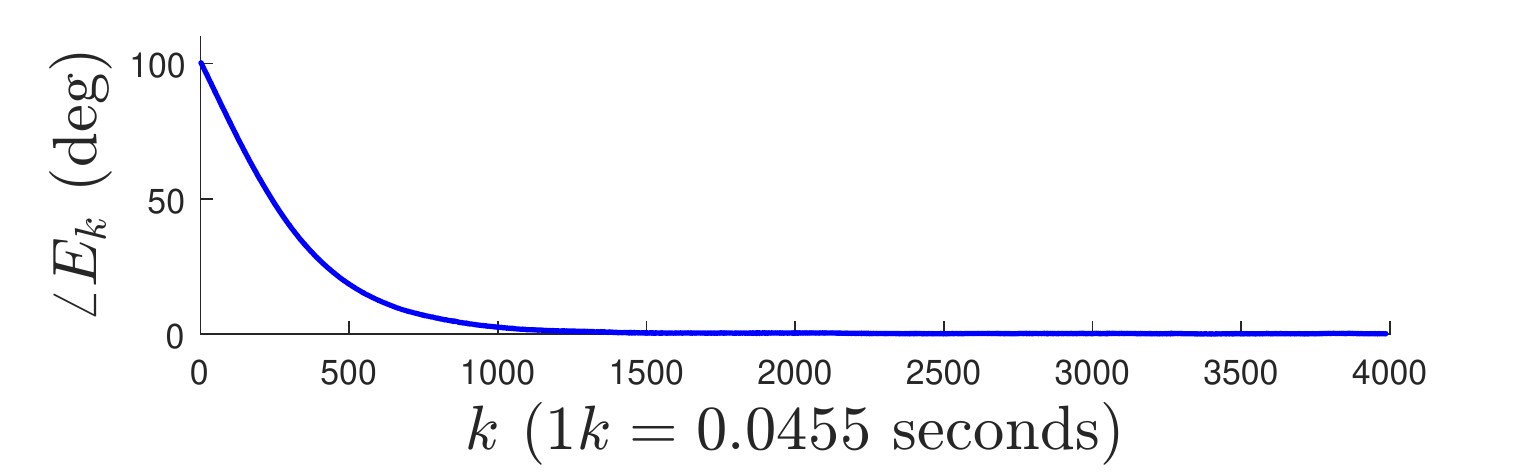}
\vspace{-3cm}

\hspace{4cm}
\includegraphics[width=4.55cm]{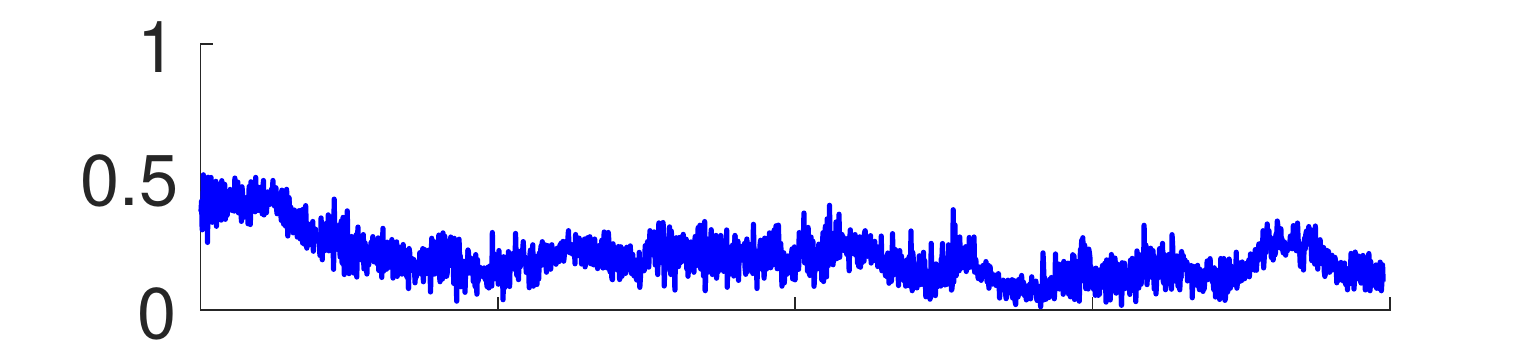}
\vspace{1cm}

\caption{Attitude estimation error. The enlarged portion of the figure shows the steady-state estimation error.} \label{fig:attitude:err:NED}
\vspace{-5mm}
\end{figure}

\begin{figure} 
\centering
\includegraphics[width=9cm,left]{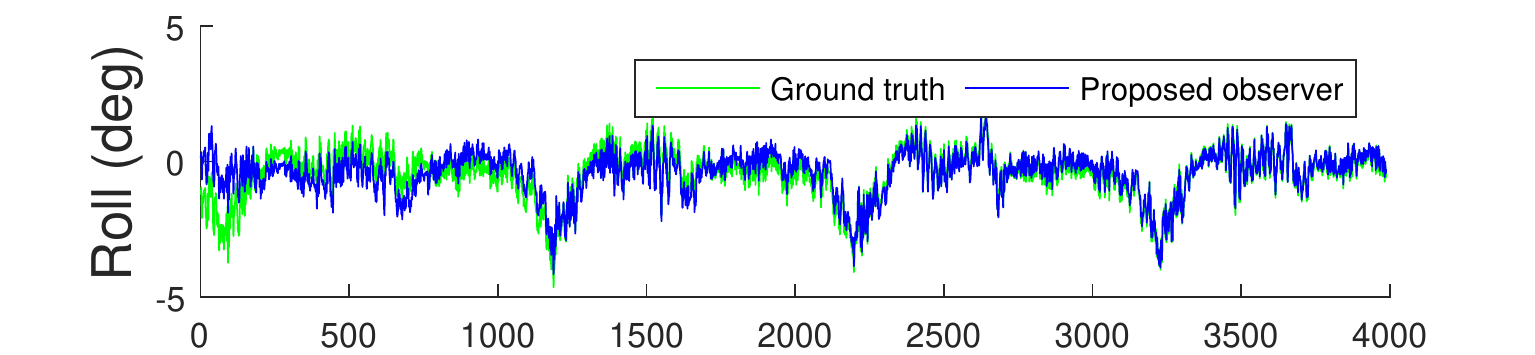}
\includegraphics[width=9cm,left]{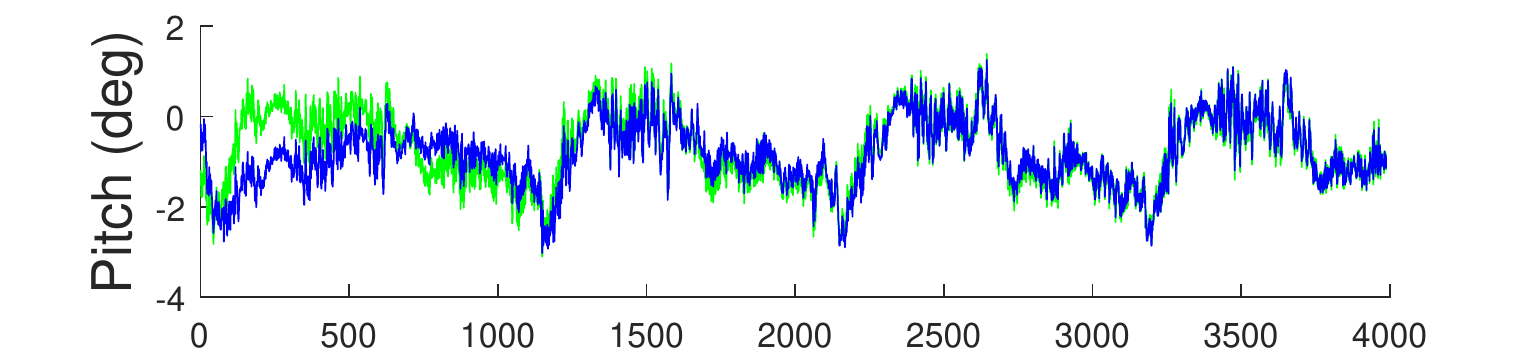}
\includegraphics[width=9cm,left]{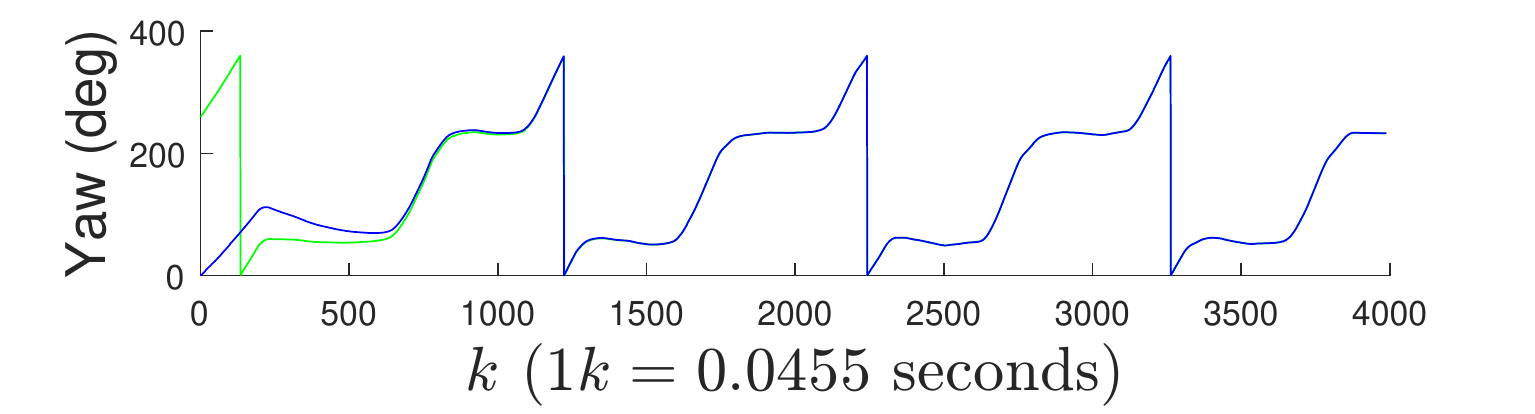}
\vspace{-0.5cm}
\caption{Ground truth Euler angle vs their estimates via the observer (\ref{eq:obs:Rk}).} \label{fig:RollPitchYaw:NED}
\vspace{-3mm}
\end{figure}

After careful calibration, the camera intrinsic parameters are estimated as $(1061.01,1062.22)$ for the focal length and $(951.28,604.86)$ for the principal point. The truck follows the path shown in Fig. \ref{fig:trajectory} during which $4000$ images are captured. Since raw velocity measurement log was not available in our experiment, we use the numerical difference of the position estimates of the INS system as estimates of the linear velocity in our observer. This yields noisy estimates of the linear velocity, but is good enough for the purpose of demonstration in this section. We choose $L=0.004 I_3$ and we initialise the observer (\ref{eq:obs:Rk}) with $\hat R_0=I_3$ since the vision-GPS system does not initially have any information about the rotation of the camera wrt. the NED frame. Fig. \ref{fig:attitude:err:NED} shows that despite the very large initial attitude estimation error, $\angle E_k$ converges to a very close neighborhood of zero. The observer is capable of accumulating the information of velocity measurements over time to asymptomatically estimate the camera's orientation wrt. the NED frame. Fig. \ref{fig:RollPitchYaw:NED} shows the Euler angle decomposition of the ground truth orientation $R_k$ versus its estimate $\hat R_k$. The Euler angles estimates follow the ground truth angles very closely, demonstrating excellent performance of the observer. The covariance of the INS attitude estimate is about $0.2$ degrees in our experiment. Fig. \ref{fig:attitude:err:NED} and Fig. \ref{fig:RollPitchYaw:NED} show that the vision-GPS system is capable of providing the level of accuracy close to a high-end INS system, albeit at a fraction of its cost.

\begin{figure}
    \centering
    \begin{subfigure}[b]{9cm}
    	\includegraphics[width=9cm]{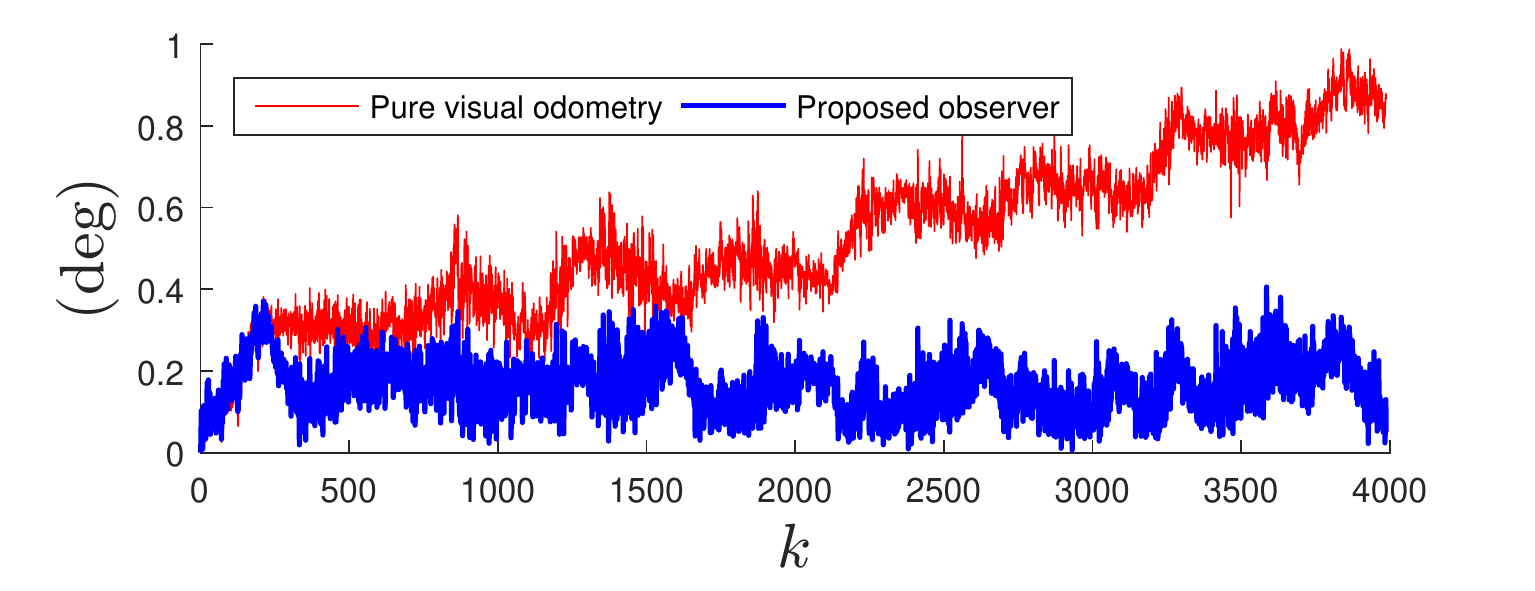}
    	\vspace{-7mm}
    	\caption{}
    	\label{fig:attitude:err:initialized}
    	\vspace{-1mm}
    \end{subfigure}
    \begin{subfigure}[b]{9cm}
    	\includegraphics[width=9cm]{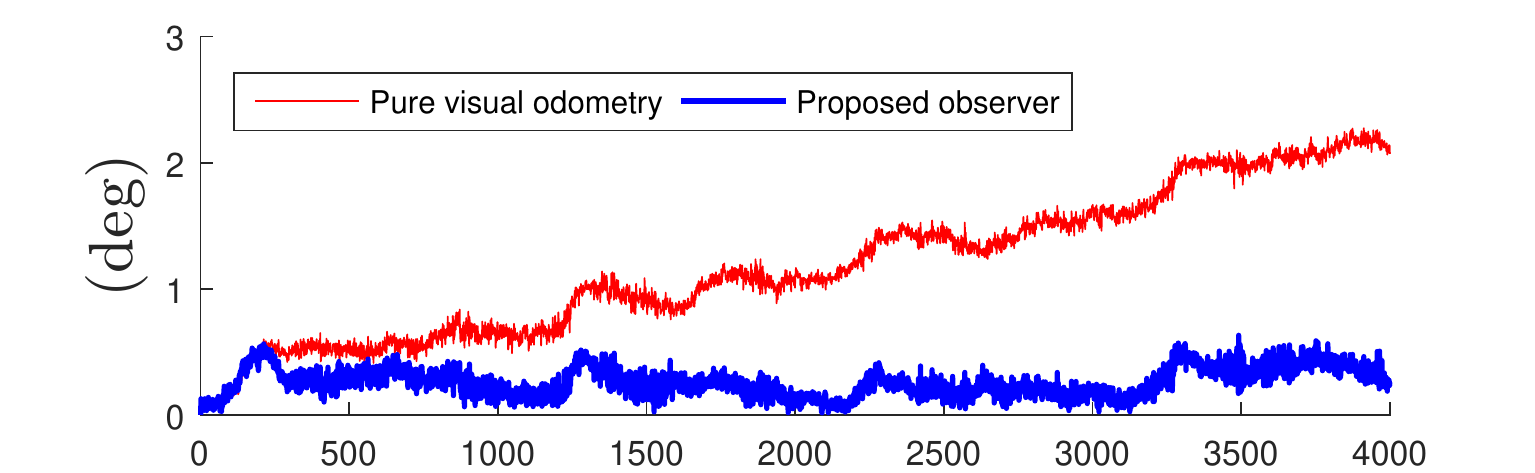}
    	\vspace{-6mm}
    	\caption{}
    	\label{fig:attitude:err:initialized:lowresolution}
    	\vspace{-1mm}
    \end{subfigure}
    \begin{subfigure}[b]{9cm}
    	\includegraphics[width=9cm]{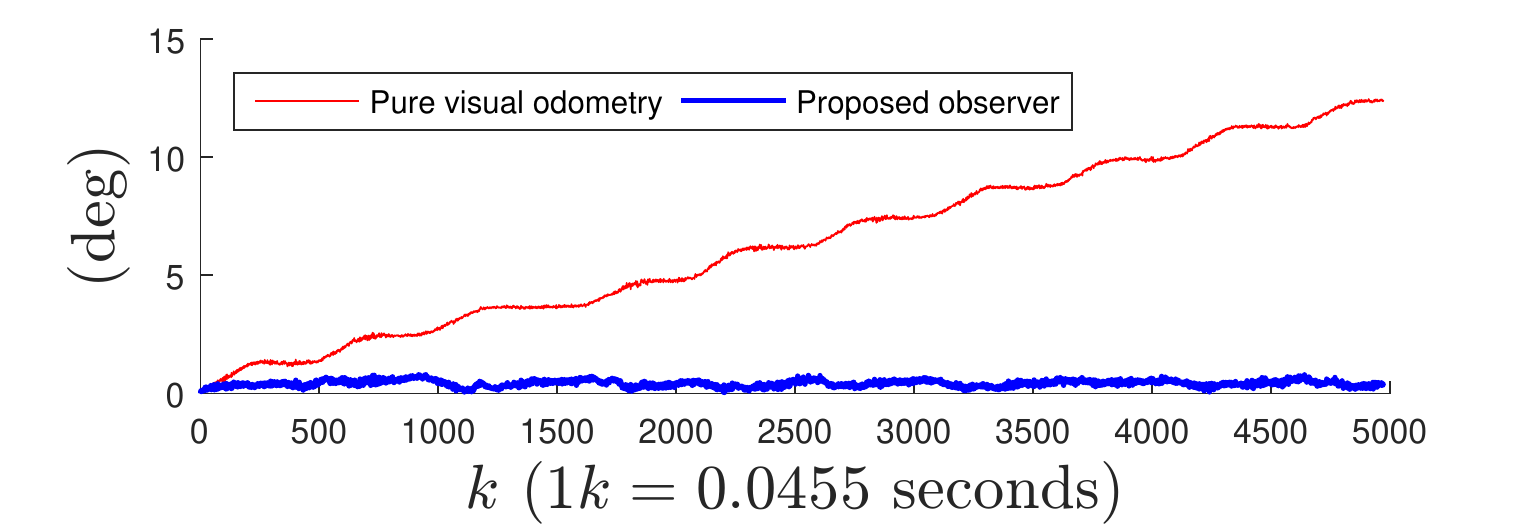}
    	\vspace{-7mm}
    	\caption{}
    	\label{fig:attitude:err:initialized:calibration}
    	\vspace{-2mm}
    \end{subfigure}
    \caption{Attitude estimation error of the pure visual odometry versus the proposed observer (\ref{eq:obs:Rk}); (a) $1920 \times 1200$ images are used, (b) $960 \times 600$ images are used, (c) $1\%$ error is added to the intrinsic camera calibration.}
    \vspace{-5mm}
\end{figure}

As explained in Section \ref{sec:problem:formulation}, one purpose of fusing GPS measurements with vision is to mitigates the inherent drift due to accumulation of the visual odometry error. In order to show this effect, we use the ground truth initial orientation $\Rca{0}$ to transform the direct orientation estimates of ORB-SLAM $\Rorb{k}{0}$ to the NED frame via $\check R_k:= \Rca{0} \Rorb{k}{0}$. This gives the estimates of $R_k$ with pure visual odometry if the initial attitude was known. We also initialise the observer (\ref{eq:obs:Rk}) with $\hat R_0=\Rca{0}$ and obtain the attitude estimates $\hat R_k$. Fig. \ref{fig:attitude:err:initialized} shows that the attitude estimation error of the pure visual odometry $\angle (\check R_k R_k^\top)$ (red curve) drifts over time while the proposed observer (\ref{eq:obs:Rk}) successfully compensates for this drift by fusing the velocity such that the magnitude of the attitude error $\angle E_k$ (blue curve) remains bounded for all times.
All of the results presented so far are obtained by processing the full resolution images. The visual odometry algorithm might not yield real time performance with full resolution of the camera. Here, we resized the images to $960 \times 600$ and perform the same experiment as Fig. \ref{fig:attitude:err:initialized}. Fig. \ref{fig:attitude:err:initialized:lowresolution} compares the attitude estimation error of the pure visual odometry versus that of the observer (\ref{eq:obs:Rk}). Compared to Fig. \ref{fig:attitude:err:initialized}, the accumulation of the visual odometry error is significantly higher while the observer error remains almost unchanged. 
Lastly, we point out that the presented results so far are obtained after very careful calibration of the camera intrinsic parameters. Even slight calibration error may significantly increase the drift of the pure visual odometry results while such drifts are compensated when the observer (\ref{eq:obs:Rk}) is employed. To demonstrate this, we add $1$ percent error to the camera intrinsic parameters used in ORB-SLAM and reprocess the full resolution images. Fig. \ref{fig:attitude:err:initialized:calibration} compares the attitude estimation of the pure visual odometry versus the observer (\ref{eq:obs:Rk}) both initialised with the ground truth orientation. Compared to Fig. \ref{fig:attitude:err:initialized}, the attitude estimation error of the pure visual odometry significantly increases while the observer (\ref{eq:obs:Rk}) successfully compensates for such a drift.

\comment{Remark: Bonnabel's paper is conservative since here we proved that for the gain l<2 the observer is still stable.}

\comment{see if we can prove the stability for a matrix gain $L$ rather than the scalar gain}

\comment{Remark: even if the VO scale $d$ drifts with time, we would yet not have any problem as we normalize.}

\comment{shall we change the matrix gain to an scalar?}

\comment{overlay the map of the experimental site on top of the vehicle trajectory}

\comment{add the point cloud of ORB-SLAM to the vehicle trajectory}

\comment{choose a lower observer gain for the down direction of the position measurement}

\comment{make a video of the truck moving around along with the estimate of vehicles trajectory and the estimation errors}

\section{Conclusion} \label{sec:conclusion}
This paper formulates the vision and GPS velocity fusion as an observer design problem on the Lie group \SO with a single vectorial output measurement whose associated reference output is time-varying. Inspired by \cite{barrau2015intrinsic,khosravian2010globally,trumpf2012analysis,grip2012attitude,Seo07}, we propose a discrete-time attitude observer and provide a rigorous stability analysis showing the exponential convergence of the attitude estimation error to zero. We present experimental studies demonstrating that the observer is effectively capable of mitigating the inherent drift of pure visual odometry estimates. The observer also enables direct attitude estimation wrt. the NED frame, making it ideal for applications involving navigation and control of robots in the inertial frame. Where depth (i.e. scale) is available through stereo vision, extending the proposed observer to a estimate the pose on SE(3) is a potential future work.


\appendices 
\section{} \label{sec:appendix:GPS2cam}
GPS measurement model (\ref{eq:GPS:difference:model}):
similar to (\ref{eq:Rc:dyn}), one can concatenate the camera positions $\pc{k+1}{0}$ to obtain
\begin{align}
\label{eq:pc:dyn} \pc{k+1}{0}-\pc{k}{0} &= \Rc{k}{0} \pc{k+1}{k},   &&\pc{0}{0}=0.
\end{align} 
On the other hand, we have
\begin{align}
\label{eq:GPS:model:tmp}	\pc{k+1}{0}-\pc{k}{0}= {\Rca{0}}^{\!\!\!\!\top} (\pca{k+1}-\pca{k}).
\end{align}
where ${\Rca{0}}^{\!\!\!\!\top}$ transforms the vector $\pca{k+1}-\pca{k}$ from the reference frame $\{A\}$ to the frame $\{C_0\}$. Substituting for (\ref{eq:GPS:model:tmp}) into (\ref{eq:pc:dyn}) and noting that GPS positions are related to $\pca{k+1}-\pca{k}$ via $\pb{k+1}-\pb{k}=d(\pca{k+1}-\pca{k})$, yields (\ref{eq:GPS:difference:model}).

\section{} \label{sec:appendix:linearization}
Linearizion of error dynamics (\ref{eq:E:dy}): using $E_{k}\approx I_3 +{\epsilon_{k}}_\times$, we have $\big(L (E_k \pau{k}-\pau{k})\!\big)_{\!\times} E_k \pau{k} \approx \big(L ((I_3 +{\epsilon_{k}}_\times) \pau{k}-\pau{k})\!\big)_{\!\times} (I_3 +{\epsilon_{k}}_\times) \pau{k} \approx \big(L {\epsilon_{k}}_\times \pau{k}\big)_{\!\times} (I_3 +{\epsilon_{k}}_\times) \pau{k} \approx \big(L {\epsilon_{k}}_\times \pau{k}\big)_{\!\times} \pau{k}$ where we ignore the terms that are of order two or higher on $\epsilon_k$. Using the Taylor series expansion of the exponential map, we have $\exp\!\Big(\Big(\!\!\big(L (E_k \pau{k}-\pau{k})\!\big)_{\!\times} E_k \pau{k}\Big)_{\!\times}\Big) E_k \approx (I_3 +  \big((L {\epsilon_{k}}_\times \pau{k})_{\!\times} \pau{k} \big)_\times + \text{HOT}) (I_3 + {\epsilon_k}_\times) \approx I_3 + {\epsilon_k}_\times + \big((L {\epsilon_{k}}_\times \pau{k})_{\!\times} \pau{k} \big)_\times +\text{HOT}$. Substituting this into (\ref{eq:E:dy}) and considering $E_{k+1} \approx I_3 + {\epsilon_{k+1}}_\times$, we have $I_3 + {\epsilon_{k+1}}_\times = I_3 + {\epsilon_k}_\times + \big((L {\epsilon_{k}}_\times \pau{k})_{\!\times} \pau{k} \big)_\times$. Canceling $I_3$ from the sides and taking the inverse of the operator $(.)_\times$ from the sides, we obtain $\epsilon_{k+1} = \epsilon_k + (L {\epsilon_{k}}_\times \pau{k})_{\!\times} \pau{k}$. Observing $(L {\epsilon_{k}}_\times \pau{k})_{\!\times} \pau{k} = \pau{k}{}_\times L \pau{k}{}_\times \epsilon_{k}$, noting $L=l I_3$, and using the property $a_\times b_\times= b a^\top - a^\top b I_3$, yields (\ref{eq:err:dy:linear}).

\section{} \label{sec:appendix:integ:inequality}
Proof of (\ref{eq:inequality:int}):  using (\ref{eq:err:dy:linear}), we have
\begin{align}
\label{eq:inequality:int:tmp1}	\epsilon_i-\epsilon_k= \sum_{j=k}^{i-1} \epsilon_{j+1}-\epsilon_j=-l \sum_{j=k}^{i-1} P_j \epsilon_j.
\end{align}
Since the maximum singular value of $P_i$ is one, and using (\ref{eq:inequality:int:tmp1}) we have
$\sum_{i=k}^{k+T} (\epsilon_i^\top - \epsilon_k^\top) P_i (\epsilon_i - \epsilon_k) \le \sum_{i=k}^{k+T} (\epsilon_i^\top - \epsilon_k^\top) (\epsilon_i - \epsilon_k) 
= l^2 \sum_{i=k}^{k+T} \big(\sum_{j=k}^{i-1} \epsilon_j^\top P_j \big)\big(  \sum_{j=k}^{i-1} P_j \epsilon_j \big)$.
Using the Cauchy–-Schwarz inequality \cite{abramowitz1964handbook} and noting $P_j P_j=P_j$, we obtain
$\sum_{i=k}^{k+T} (\epsilon_i^\top - \epsilon_k^\top) P_i (\epsilon_i - \epsilon_k) \le
	l^2 \sum_{i=k}^{k+T}  (i-k) \sum_{j=k}^{i-1} \epsilon_j^\top P_j  P_j \epsilon_j
	=l^2 \sum_{i=k+1}^{k+T} (i-k) \sum_{j=k}^{i-1} \epsilon_j^\top P_j \epsilon_j,
$
where we used the fact that the value of the first summand evaluated at $i=k$ is zero and hence we can start the first summation from $i=k+1$. Changing the order of summations we obtain
\begin{align}
	\label{eq:inequality:integral:tmp2} \sum_{i=k}^{k+T} &(\epsilon_i^\top - \epsilon_k^\top) P_i (\epsilon_i - \epsilon_k) \le 
		l^2 \sum_{j=k}^{k+T-1} \epsilon_j^\top P_j \epsilon_j \sum_{i=j+1}^{k+T} (i-k).  
\end{align}
We have $\sum_{i=j+1}^{k+T} (i-k) = \frac{T(T+1)}{2}-\frac{(j-k+1)(j-k)}{2}$. Using the limits of the first summation we have $k \le j \le k+T-1$ which implies $0 \le (j-k+1)(j-k) \le T(T-1)$. Hence $\sum_{i=j+1}^{k+T} (i-k) \le  \frac{T(T+1)}{2}$. Using this and adding the positive term $\frac{l^2T(T+1)}{2} \epsilon_{k+T}^\top P_{k+T} \epsilon_{k+T}$ to the right hand side of (\ref{eq:inequality:integral:tmp2}) yields  
(\ref{eq:inequality:int}) and completes the proof.

\bibliographystyle{IEEEtran}
\bibliography{librarysample}

\begin{thebibliography}{10}
\providecommand{\url}[1]{#1}
\csname url@samestyle\endcsname
\providecommand{\newblock}{\relax}
\providecommand{\bibinfo}[2]{#2}
\providecommand{\BIBentrySTDinterwordspacing}{\spaceskip=0pt\relax}
\providecommand{\BIBentryALTinterwordstretchfactor}{4}
\providecommand{\BIBentryALTinterwordspacing}{\spaceskip=\fontdimen2\font plus
\BIBentryALTinterwordstretchfactor\fontdimen3\font minus
  \fontdimen4\font\relax}
\providecommand{\BIBforeignlanguage}[2]{{%
\expandafter\ifx\csname l@#1\endcsname\relax
\typeout{** WARNING: IEEEtran.bst: No hyphenation pattern has been}%
\typeout{** loaded for the language `#1'. Using the pattern for}%
\typeout{** the default language instead.}%
\else
\language=\csname l@#1\endcsname
\fi
#2}}
\providecommand{\BIBdecl}{\relax}
\BIBdecl

\bibitem{markley2014fundamentals}
F.~L. Markley and J.~L. Crassidis, \emph{Fundamentals of Spacecraft Attitude
  Determination and Control}.\hskip 1em plus 0.5em minus 0.4em\relax Springer,
  2014.

\bibitem{crassidis2007survey}
J.~L. Crassidis, F.~L. Markley, and Y.~Cheng, ``Survey of nonlinear attitude
  estimation methods,'' \emph{Journal of guidance, control, and dynamics},
  vol.~30, no.~1, pp. 12--28, 2007.

\bibitem{Mahony08}
R.~Mahony, T.~Hamel, and {J.M. Pflimlin}, ``Nonlinear complementary filters on
  the special orthogonal group,'' \emph{{IEEE} Trans. Autom. Control}, vol.~53,
  no.~5, pp. 1203--1218, 2008.

\bibitem{rehbinder2003pose}
H.~Rehbinder and B.~K. Ghosh, ``Pose estimation using line-based dynamic vision
  and inertial sensors,'' \emph{{IEEE} Trans. Automatic Control}, vol.~48,
  no.~2, pp. 186--199, 2003.

\bibitem{grip2012attitude}
H.~F. Grip, T.~I. Fossen, T.~A. Johansen, and A.~Saberi, ``Attitude estimation
  using biased gyro and vector measurements with time-varying reference
  vectors,'' \emph{IEEE Trans. Autom. Control}, vol.~57, no.~5, pp. 1332--1338,
  2012.

\bibitem{bonnabel2006non}
S.~Bonnabel, P.~Martin, and P.~Rouchon, ``A non-linear symmetry-preserving
  observer for velocity-aided inertial navigation,'' in \emph{Proc. American
  Control Conf.}, 2006, pp. 2910--2914.

\bibitem{Vasconcelos08a}
J.~Vasconcelos, C.~Silvestre, and P.~Oliveira, ``A nonlinear observer for rigid
  body attitude estimation using vector observations,'' in \emph{Proc. IFAC
  World Congr.}, Korea, July 2008.

\bibitem{zlotnik2015nonlinear}
D.~E. Zlotnik and J.~R. Forbes, ``A nonlinear attitude estimator with desirable
  convergence properties,'' in \emph{European Control Conference}, 2015, pp.
  2103--2107.

\bibitem{Seo07}
D.~Seo and {M.R. Akella}, ``Separation property for the rigid-body attitude
  tracking control problem,'' \emph{J. Guid., Control, Dynam.}, vol.~30, no.~6,
  pp. 1569--1576, 2007.

\bibitem{Tayebi07McGilvray}
A.~Tayebi, S.~McGilvray, A.~Roberts, and M.~Moallem, ``Attitude estimation and
  stabilization of a rigid body using low-cost sensors,'' in \emph{Proc. {IEEE}
  Conf. on Decision and Control}, USA, December 2007.

\bibitem{fraundorfer2011visual}
F.~Fraundorfer and D.~Scaramuzza, ``Visual odometry: Part {I}: The first 30
  years and fundamentals,'' \emph{IEEE Robotics and Automation Magazine},
  vol.~18, no.~4, pp. 80--92, 2011.

\bibitem{fraundorfer2012visual}
------, ``Visual odometry: Part {II}: Matching, robustness, optimization, and
  applications,'' \emph{{IEEE} Robotics \& Automation Magazine}, vol.~19,
  no.~2, pp. 78--90, 2012.

\bibitem{murTRO2015}
R.~Mur-Artal, J.~Montiel, and J.~D. Tard{\'o}s, ``{ORB-SLAM}: a versatile and
  accurate monocular {SLAM} system,'' \emph{IEEE Trans. Robotics}, vol.~31,
  no.~5, pp. 1147--1163, 2015.

\bibitem{engel2013semi}
J.~Engel, J.~Sturm, and D.~Cremers, ``Semi-dense visual odometry for a
  monocular camera,'' in \emph{Proc. {IEEE} International Conf. Computer
  Vision}, 2013, pp. 1449--1456.

\bibitem{pireIROS15}
T.~Pire, T.~Fischer, J.~Civera, P.~De~Crist{\'oforis}, and J.~Jacobo~berlles,
  ``{Stereo Parallel Tracking and Mapping for robot localization},'' in
  \emph{Proc. {IEEE/RSJ} International Conf. Intelligent Robots and Systems},
  2015, pp. 1373--1378.

\bibitem{triggs1999bundle}
B.~Triggs, P.~F. McLauchlan, R.~I. Hartley, and A.~W. Fitzgibbon, ``Bundle
  adjustment \textemdash a modern synthesis,'' in \emph{International workshop
  on vision algorithms}, 1999, pp. 298--372.

\bibitem{angeli2008fast}
A.~Angeli, D.~Filliat, S.~Doncieux, and J.-A. Meyer, ``Fast and incremental
  method for loop-closure detection using bags of visual words,'' \emph{IEEE
  Trans. Robotics}, vol.~24, no.~5, pp. 1027--1037, 2008.

\bibitem{lefferts1982kalman}
E.~J. Lefferts, F.~L. Markley, and M.~D. Shuster, ``{Kalman} filtering for
  spacecraft attitude estimation,'' \emph{Journal of Guidance, Control, and
  Dynamics}, vol.~5, no.~5, pp. 417--429, 1982.

\bibitem{izadi2015comparison}
M.~Izadi, E.~Samiei, A.~K. Sanyal, and V.~Kumar, ``Comparison of an attitude
  estimator based on the lagrange-d'alembert principle with some
  state-of-the-art filters,'' in \emph{{IEEE} International Conf. Robotics and
  Automation}, 2015, pp. 2848--2853.

\bibitem{roberts2003low}
J.~M. Roberts, P.~I. Corke, and G.~Buskey, ``Low-cost flight control system for
  a small autonomous helicopter,'' in \emph{Proc. {IEEE} International Conf.
  Robotics and Automation}, vol.~1, 2003, pp. 546--551.

\bibitem{zamani2013deterministic}
M.~Zamani, ``Deterministic attitude and pose filtering, an embedded {Lie}
  groups approach,'' Ph.D. dissertation, Australian National University, 2013.

\bibitem{khosravian2010globally}
A.~Khosravian and M.~Namvar, ``Globally exponential estimation of satellite
  attitude using a single vector measurement and gyro,'' in \emph{Proc. {IEEE}
  Conf. Decision and Control}, 2010.

\bibitem{barrau2015invariant}
A.~Barrau \emph{et~al.}, ``Invariant filtering for pose {EKF-SLAM} aided by an
  imu,'' in \emph{IEEE Conference on Decision and Control}, 2015, pp.
  2133--2138.

\bibitem{khosravian2015observers}
A.~Khosravian, J.~Trumpf, R.~Mahony, and C.~Lageman, ``Observers for invariant
  systems on lie groups with biased input measurements and homogeneous
  outputs,'' \emph{Automatica}, vol.~55, pp. 19--26, 2015.

\bibitem{lageman2010gradient}
C.~Lageman, J.~Trumpf, and R.~Mahony, ``Gradient-like observers for invariant
  dynamics on a {Lie} group,'' \emph{IEEE Trans. Automa. Control}, vol.~55,
  no.~2, pp. 367--377, 2010.

\bibitem{bonnabel2009non}
S.~Bonnabel, P.~Martin, and P.~Rouchon, ``Non-linear symmetry-preserving
  observers on {Lie} groups,'' \emph{IEEE Trans. Autom. Control}, vol.~54,
  no.~7, pp. 1709--1713, 2009.

\bibitem{barrau2015intrinsic}
A.~Barrau and S.~Bonnabel, ``Intrinsic filtering on lie groups with
  applications to attitude estimation,'' \emph{IEEE Transactions on Automatic
  Control}, vol.~60, no.~2, pp. 436--449, 2015.

\bibitem{lee07}
T.~Lee, M.~Leok, N.~H. McClamroch, and A.~Sanyal, ``Global attitude estimation
  using single direction measurements,'' in \emph{Proc. American Control
  Conf.}, USA, July 2007.

\bibitem{trumpf2012analysis}
J.~Trumpf, R.~Mahony, T.~Hamel, and C.~Lageman, ``Analysis of non-linear
  attitude observers for time-varying reference measurements,'' \emph{IEEE
  Trans. Autom. Control}, vol.~57, no.~11, pp. 2789--2800, 2012.

\bibitem{hartley2003multiple}
R.~Hartley and A.~Zisserman, \emph{Multiple view geometry in computer
  vision}.\hskip 1em plus 0.5em minus 0.4em\relax Cambridge university press,
  2003.

\bibitem{mahony2013observers}
R.~Mahony, J.~Trumpf, and T.~Hamel, ``Observers for kinematic systems with
  symmetry,'' in \emph{Proc. {IFAC} Symposium on Nonlinear Control Systems},
  September 2013, pp. 617--633.

\bibitem{khosravian2013bias}
A.~Khosravian, J.~Trumpf, R.~Mahony, and C.~Lageman, ``Bias estimation for
  invariant systems on {Lie} groups with homogeneous outputs,'' in \emph{Proc.
  IEEE Conf. on Decision and Control}, December 2013.

\bibitem{hua2010attitude}
M.-D. Hua, ``Attitude estimation for accelerated vehicles using {GPS/INS}
  measurements,'' \emph{Control Engineering Practice}, vol.~18, no.~7, pp.
  723--732, 2010.

\bibitem{kalman1960control}
R.~Kalman and J.~Bertram, ``Control system analysis and design via the second
  method of lyapunov: Ii discrete-time systems,'' \emph{Journal of Basic
  Engineering}, vol.~82, no.~2, pp. 394--400, 1960.

\bibitem{Khalil}
H.~K. Khalil, \emph{Nonlinear Systems}, 3rd~ed.\hskip 1em plus 0.5em minus
  0.4em\relax Prentice Hall, 2002.

\bibitem{horaud1995hand}
R.~Horaud and F.~Dornaika, ``Hand-eye calibration,'' \emph{The international
  journal of robotics research}, vol.~14, no.~3, pp. 195--210, 1995.

\bibitem{abramowitz1964handbook}
M.~Abramowitz and I.~A. Stegun, \emph{Handbook of mathematical functions: with
  formulas, graphs, and mathematical tables}.\hskip 1em plus 0.5em minus
  0.4em\relax Courier Corporation, 1964, vol.~55.

\end{thebibliography}

\end{document}